\documentclass{article}

\usepackage{PRIMEarxiv}

\usepackage[utf8]{inputenc} 
\usepackage[T1]{fontenc}    
\usepackage{hyperref}       
\usepackage{url}            
\usepackage{booktabs}       
\usepackage{amsfonts}       
\usepackage{nicefrac}       
\usepackage{microtype}      
\usepackage{lipsum}
\usepackage{fancyhdr}       
\usepackage{graphicx}       
\graphicspath{{media/}}     
\usepackage{amssymb}
\usepackage{multirow}
\usepackage{array}
\usepackage[T1]{fontenc}
\usepackage[utf8]{inputenc}
\usepackage{times}

\title{Extracting Abstraction Dimensions by Identifying Syntax Pattern from Texts
}

\author{
  Jian Zhou, Jiazheng Li \\
  Key Laboratory of Intelligent Information Processing\\
  Institute of Computing Technology, Chinese Academy of Sciences, Beijing, China\\
  University of Chinese Academy of Sciences, Beijing, China\\
  \texttt{\{newzhoujian, leejiazh6\}@gmail.com} \\
   \And
  Sirui Zhuge \\
  King’s College London, UK \\
  Publicis Sapient, UK \\
  \texttt{sirui.zhuge@kcl.ac.uk} \\
   \And
  Hai Zhuge\thanks{Hai Zhuge is the corresponding author} \\
  School of Computing and Information Technology, Great Bay University, Dongguan, China \\
 Great Bay Institute for Advanced Study, Dongguan, China  \\
  \texttt{zhuge@gbu.edu.cn} \\
}

\begin{document}
\maketitle

\begin{abstract}
This paper proposed an approach to automatically discovering subject dimension, action dimension, object dimension and adverbial dimension from texts to efficiently operate texts and support query in natural language.  The high quality of trees guarantees that all subjects, actions, objects and adverbials and their subclass relations within texts can be represented. The independency of trees ensures that there is no redundant representation between trees. The expressiveness of trees ensures that the majority of sentences can be accessed from each tree and the rest of sentences can be accessed from at least one tree so that the tree-based search mechanism can support querying in natural language.  Experiments show that the average precision, recall and F1-score of the abstraction trees constructed by the subclass relations of subject, action, object and adverbial are all greater than 80\%.  The application of the proposed approach to supporting query in natural language demonstrates that different types of question patterns for querying subject or object have high coverage of texts, and searching multiple trees on subject, action, object and adverbial according to the question pattern can quickly reduce search space to locate target sentences, which can support precise operation on texts.
\end{abstract}


\section{Introduction}
The abstraction dimension can be understood as a classification method in the form of a class tree composed of classes and the subclass relations between classes.  For a set of texts, different classification methods result in different class trees, thereby forming different abstraction dimensions that can locate the texts efficiently from multiple dimensions.  The abstraction dimension is different from the dimensions in the Space Vector Model (SVM), which regards the non-duplicate terms of a set of texts as dimensions.

Discovering abstraction dimensions from texts is a way to efficiently support advanced services, e.g., a question-answer system can input questions in natural language and then outputs answers in the same language. Using a space of multiple abstraction dimensions to manage texts enables a small set of texts (as an answer) to be quickly accessed from different dimensions \cite{zhuge2016multi}.

The Resource Space Model (RSM) is a classification-based resource management model that organizes, specifies, and manages resources with a resource space consisting of multiple abstraction dimensions that represent subclass relations between various resources.  Each subclass represents fewer resources of its father class (superclass) as a superclass represents a more general class \cite{zhuge2016multi,zhuge2008web,zhuge2012knowledge,zhuge2011probabilistic,zhuge2008resource}.

For text resources, the sentence represents the complete meaning of the event and state of the thing, so it is the main form of query and answer.  A syntax pattern is a kind of sentence abstraction.   The basic syntax pattern of sentences consists of subject, action, object, and adverbial, which represent four dimensions of sentences, i.e., sentences within texts can be commonly viewed or accessed from subject, action, object, and adverbial.

\subsection{Motivation Example}
The major question-answering approaches including ranking-based approaches and Large Language Model (LLM)-based approach are compared by inputting a short text and a long text in Table \ref{tab:table1}.

The ranking-based approaches (also called bag-of-words approaches including Common Words, Jaccard Similarity, TFIDF+Cosine Similarity, Word Embedding+Cosine Similarity, Unigram Language Model and Okapi BM25) measure the relevance between question and candidate answer based on the number of common words, the weights (term frequency, inverse document frequency, or the conditional probability of a word appearing) of common words, or cosine similarity of word embedding vectors.  For short input texts, these approaches select Sentence 3 as the answer because it has the most common words with the question, but it is an incorrect answer to the question (labelled as “$\times$”).  For long input texts, the approaches based on the number of common words and the weights of common words top rank sentence 29 “The CREF (Cross-Referenced Editing Facility) system supported four kinds of link: the reference link for cross-referring, the summarization link for summarizing, the supersede link for versioning, and the precede link for ordering” which shares more common words with the question than other sentences. However, the approach based on cosine similarity between embedding vectors of sentences and question top ranks sentence 208 “Web of things can be regarded as a kind of cyber-physical space” with similarity 0.96, and ranks sentence 29 with similarity 0.88 as it contains more words that do not appear in the question than sentence 208. Sentence 29 is the correct answer (labelled as “$\checkmark$” ), and sentence 208 is an incorrect answer. Ranking-based approaches based on word sequence (such as Greedy String Tiling and Longest Common Subsequence) measure the relevance of the question and candidate answer by using the longest common words sequence.  For short input texts, these approaches select sentence 4 as the answer as it has the top common words sequence with the question, i.e., “algorithm”, “build” and “extract”, but it is a wrong answer.  For long input texts, these approaches select sentence 29 as it has the longest common-word sequence with the question compared to other sentences.

LLM-based approaches generate answers by predicting the next word according to a language model.   Herein, ChatGPT is tested by inputting the following prompt: “Read the following texts: <Input Texts>, and answer the following question based on the above texts. The question is <Question> and the answer is limited within one sentence”.   For short input texts, the answer of ChatGPT is not faithful to the input text as the part “based on their similarity or proximity to each other in a multidimensional space.” is not consistent with the input.  For long input texts, ChatGPT fabricates the answer. The long input paper in Table \ref{tab:table1} does mention “single-type links” and “complex links”, but the relations between “CREF system” and “single-type links”, “complex links” is not mentioned, and sentence 29 introduces the “CREF system” can support “the reference link”, “the summarization link”, “the supersede link” and “the precede link”. As being trained on a large-scale dataset to generate answers according to conditional probability without querying a database of facts, LLM as ChatGPT may lead to factual errors.  Besides, ChatGPT has the input length limitation which makes it difficult to answer based on large-scale input texts.

\begin{figure}
  \centering
  \includegraphics[width=0.9\linewidth]{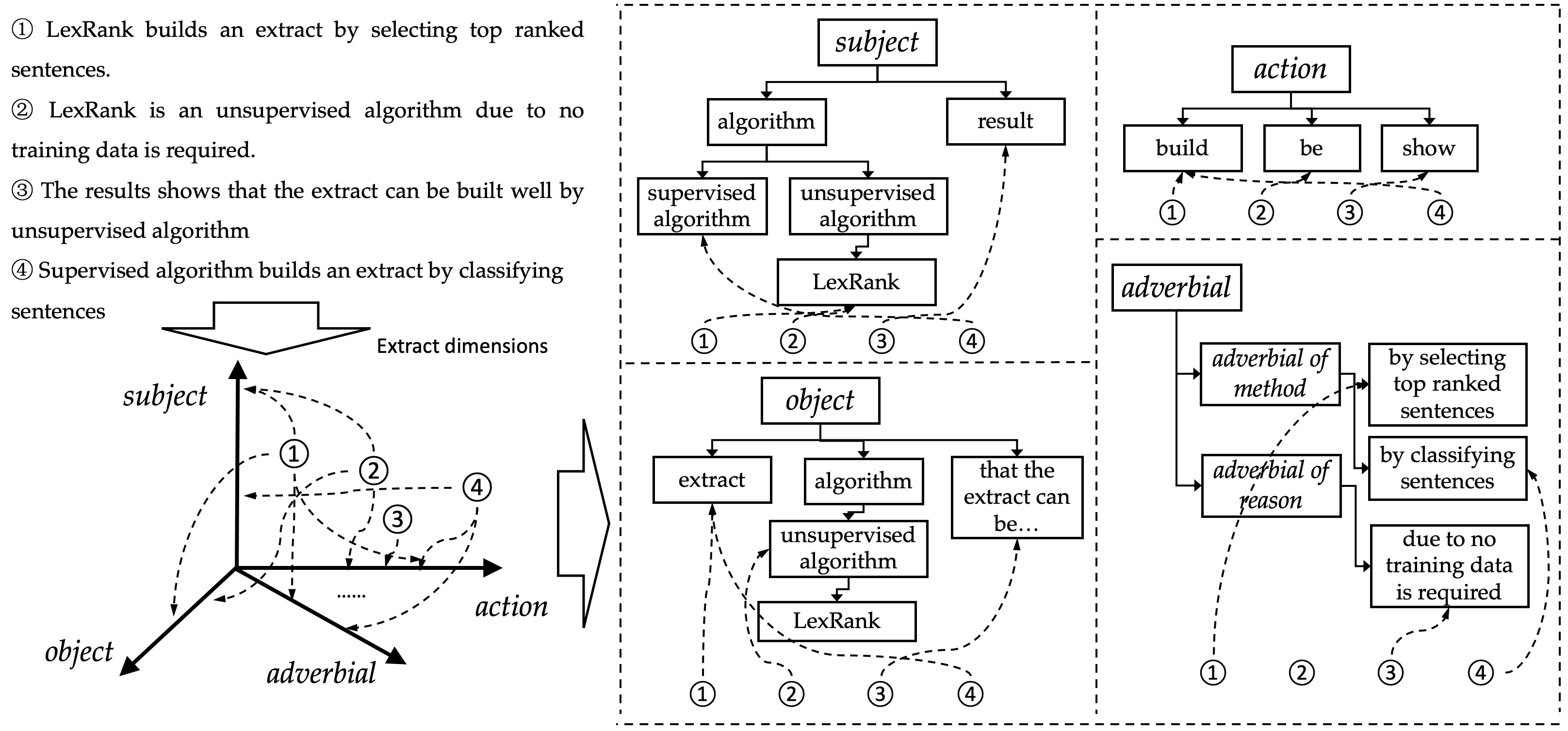}
  \caption{Four syntax dimensions extracted from the short input texts.}
  \label{fig:fig1}
\end{figure}

Discovering the dimensions of input texts is a way to efficiently support question answering. As shown in Figure \ref{fig:fig1}, a resource space composed of four syntax dimensions (subject, action, object, and adverbial) can be discovered by extracting the subjects, actions, objects, and adverbials from the sentences in the texts to be managed and by identifying their subclass relations. The sentences identified by the subclass can also be identified by the its superclass due to the characteristic of the subclass relations, e.g., in the subject dimension, the “LexRank” is the subclass of the “unsupervised algorithm” (identified by the syntactic pattern “$NP_{hyper}$ is an $NP_{hypon}$” from Sentence 2, where $NP_{hyper}$ represents the superclass noun phrase, and $NP_{hypon}$ represents the subclass noun phrase). When accessing the sentences identified by the class “unsupervised algorithm” in the subject dimension, Sentences 1 and 2 can also be accessed. Based on the extracted subject, action, object, and adverbial of the question of the input texts (i.e., “unsupervised algorithm” as the subject can locate Sentences 1 and 2, “build” as the action can locate Sentences 1 and 4, “extract” as the object can locate Sentences 1 and 4), only Sentence 1 is selected as the candidate answer. As Sentence 1 has the adverbial of the method (i.e., “by selecting top ranked sentences”) that does not appear in the question, it can be used to answer the how question. Therefore, Sentence 1 is selected as the answer from the multi-dimensional space.

\begin{table}
  \caption{Ranking-based approaches and LLM-based approach in answering questions.}
  \centering
  \begin{tabular}{>{\raggedright\arraybackslash}p{2cm} >{\raggedright\arraybackslash}p{3cm} >{\raggedright\arraybackslash}p{6cm} >{\raggedright\arraybackslash}p{5cm}}
    \toprule
    \textbf{Types of Methods} & \textbf{Methods} & \textbf{Short Input Texts} & \textbf{Long Input Texts} \\
    \cmidrule(r){3-4}
     & & \textbf{Input Texts:} four sentences as follows:\newline
1. LexRank builds an extract by selecting top ranked sentences.\newline
2. LexRank is an unsupervised algorithm due to no training data is required.\newline
3. The results show that the extract can be built well by unsupervised algorithm.\newline
4. Supervised algorithm builds an extract by classifying sentences.
     & \textbf{Input Texts:} The scientific literature “Semantic linking through spaces for cyber-physical-socio intelligence: A methodology”, about 1077 sentences and 18795 words. \\
    \cmidrule(r){3-4}
    & & \textbf{Question}: How does unsupervised algorithm build an extract? & \textbf{Question}: What kinds of link does CREF (Cross-Referenced Editing Facility) system support? \\
    \midrule
    \multirow{8}{2cm}{Ranking-based Approach} & Common Words & Sentence 3 ($\times$) & Sentence 29 ($\checkmark$) \\ 
    \cmidrule(r){2-4}
     & Jaccard Similarity & Sentence 3 ($\times$) & Sentence 29 ($\checkmark$) \\ 
     \cmidrule(r){2-4}
     & TFIDF + Cosine Similarity & Sentence 3 ($\times$) & Sentence 29 ($\checkmark$) \\ 
     \cmidrule(r){2-4}
     & Word Embedding + Cosine Similarity & Sentence 3 ($\times$) & Sentence 208 ($\times$) \\ 
     \cmidrule(r){2-4}
     & Unigram Language Models & Sentence 3 ($\times$) & Sentence 29 ($\checkmark$) \\ 
     \cmidrule(r){2-4}
     & Okapi BM25 & Sentence 3 ($\times$) & Sentence 29 ($\checkmark$) \\ 
     \cmidrule(r){2-4}
     & Greedy String Tiling (GST) & Sentence 4 ($\times$) & Sentence 29 ($\checkmark$) \\ 
     \cmidrule(r){2-4}
     & Longest Common Subsequences (LCS) & Sentence 4 ($\times$) & Sentence 29 ($\checkmark$) \\ 
     \midrule
     LLM-based Approach & ChatGPT & An unsupervised algorithm builds an extract by selecting top ranked sentences based on their similarity or proximity to each other in a multidimensional space. ($\times$) & The CREF (Cross-Referenced Editing Facility) system supports both single-type links and complex links in its cyber-physical-socio environment. ($\times$) \\
    \bottomrule
  \end{tabular}
  \label{tab:table1}
\end{table}

This example shows that (1) ranking-based approaches may find wrong answer because the sentence sharing the most number of words with question may not answer the question because common words may not represent the purpose of question; (2) LLM-based approaches may generate answers that are not faithful to the original text; and (3) extracting subject dimension, action dimension, object dimension, and adverbial dimension can find sentences that answer questions about subject, action, object, and adverbial, which cover all questions.  Therefore, it is necessary to extract abstraction dimensions from the texts.

\subsection{General Idea for Extracting Subclass Relations}
Subject, action, object, and adverbial are the four syntax elements for describing things.  A subject refers to a thing that performs an action represented by verb or verb phrases.   An object refers to a thing that receives an action.  An adverbial modifies an action and provides additional information about the time, place, method, purpose, reason or condition of the action.

Sentences can be classified according to their subjects, actions, objects or adverbials as most sentences contain these syntax elements.  Sentences with shared subjects, actions, objects, or adverbials reflect commonalities among them.  People can better understand events by identifying actions, performers of actions, receivers of actions, and circumstances of when, where, how, and why actions occur.

The proposed approach is to automatically discover the abstraction dimensions of the subject, action, object, and adverbial from a set of texts and supports precise query applications.  The key problem is to identify the subject, action, object, and adverbial of sentences and their subclass relations respectively.

The first issue is to \textit{identify the language representations of syntax elements}.

 English uses a combination of subject, action, object, and adverbial to represent a sentence. Identification of subject, action, object, and adverbial is determined by the structure of the combination and part-of-speech of words that construct them.  For example, the subject is determined by the order of combination (it appears before the action) and whether it is represented by a noun or a nominal representation.
 
The second issue is \textit{how to identify the subclass relations between subjects, between actions, between objects and between adverbials respectively}.

As different types of phrases belong to different lexical categories, subclass relation exists between the phrases that have the same lexical category, e.g., subclass relation exists between noun phrases rather than between noun phrase and verb phrase.

There are three approaches to identifying subclass relations between phrases: 

\begin{enumerate}
\item \textit{Syntactic approach}, i.e., using empirical syntactic patterns to represent subclass relations between phrases.  For noun phrases, “$NP_{hyper}$ including $NP_{hypon}$” is a pattern representing subclass relations between a class (i.e., $NP_{hyper}$) and its subclass (i.e., $NP_{hypon}$) \cite{hearst1992automatic}. For verbs, “to $VP_{hypon}$ is to $VP_{hyper}$” is a pattern representing a subclass relations between the class verb (i.e., $VP_{hyper}$) and its subclass verb (i.e., $VP_{hypon}$) \cite{huminski2018wordnet}.
\item \textit{Modifier approach}, i.e., using words to restrict the meaning of a word or a phrase \cite{zhou2025automatically}. It is suitable for the phrase composed of pre-head, head and post-head such as noun phrase (e.g., “extractive summarization” as the subclass of “summarization”), verb phrase (e.g., “go slowly” as the subclass of “go”), and adjective phrase (e.g., “very beautiful” as the subclass of “beautiful”).
\item \textit{Integrated approach, that incorporates the syntactic approach and the modifier approach}.  For phrases such as the prepositional phrase that are not composed of pre-head, head and post-head, the subclass relation between phrases can be identified by the following two steps: 1) identifying whether there is a subclass relation between the parts that consist of the pre-head, head and post-head of the two phrases by the modifier approach or the syntactic approach, and 2) judging whether the rest of the two phrases are the same or not.  For example, “with text summarization technique” is the subclass of “with summarization technique”, and “based on LexRank” is the subclass of “based on unsupervised algorithm” if “LexRank” is identified as the subclass of “unsupervised algorithm” by syntactic approach.
\end{enumerate}

A clause is a group of words consists of lead words (i.e., subordinate conjunctions in subordinating clause and “to” in infinitive clause), subject, action, object, and adverbial. The subclass relation between clauses can be identified by judging 1) the equality of lead words of the clauses, and 2) the subclass relations between the phrases representing the subject, action, object, and adverbial of the clauses.  For example, the clause “what unsupervised algorithm can do” is the subclass of “what algorithm can do”.

\subsection{General Process for Extracting Dimension and Query}
The general process consists of the following steps:
\begin{enumerate}
\item \textit{Extracting dimensions from input texts}. It consists of the following steps: 1) Identifying the patterns of the subject, action, object, and adverbial. 2) Extracting the subject, action, object, and adverbial of each sentence from the input texts according to patterns. 3) Identifying the subclass relations between subjects, between actions, between objects and between adverbials respectively to construct abstraction trees. As shown in Figure 2, the subject “spreading activation algorithm” within text (1) can be viewed as the subclass of the subject “spreading algorithm” within text (2) by the modifier approach. Within text (8), the noun “LexRank” can be viewed as the subclass of the noun phrase “unsupervised algorithm” by syntactic pattern “$NP_{hypon}$ is an $NP_{hyper}$”. 4) Combining the same language representation of subject, action, object, and adverbial and removing the subclass relations that can be derived from the existing subclass relations to make simpler abstraction trees that represent the same semantics.

\item \textit{Verifying whether the extracted abstraction trees construct dimensions}. It carries out with judging: 1) the quality (measured by precision, recall and F1-score) of the subclass relations of the extracted the subjects, actions, objects, and adverbials; 2) the independence criteria (dimensions are independent of each other, i.e., there are no identical nodes because a node identifies a class of resources) and the expressiveness criteria (dimensions have a high coverage of texts, i.e., dimensions have the ability to represent or access most of the texts, which is the basis of searching the texts and being able to support question answering).

\item \textit{Processing query}. Processing questions in natural language consists of the following steps:  1) Identify the patterns of questions and answers.  2) Parse and extract subject, action, object, and adverbial of question according to patterns.  As Q2 shown in Figure 2, the subject is the “graph-based unsupervised algorithm”, the action is “select” (“does” is the auxiliary and can be omitted), object is empty (as it is asked by the question) and the adverbial is purpose adverbial “to composed an extract”.  3) Search the dimensions of the resource space with the extracted subject, action, object, and adverbial and then return the matched sentences as the candidate answers.  4) Select the answers that contain the syntax elements (i.e., subject, action, object, and adverbial) that the question asks for.  For example, the answer to the when question should include adverbial on time.
\end{enumerate}

The resource space composed of dimensions of subject, action, object, and adverbial can effectively support queries in natural language. It can obtain answers (a small set of sentences) efficiently by searching multiple dimensions according to identification of subject, action, object, and adverbial within natural language questions.

\begin{figure}
  \centering
  \includegraphics[width=1\linewidth]{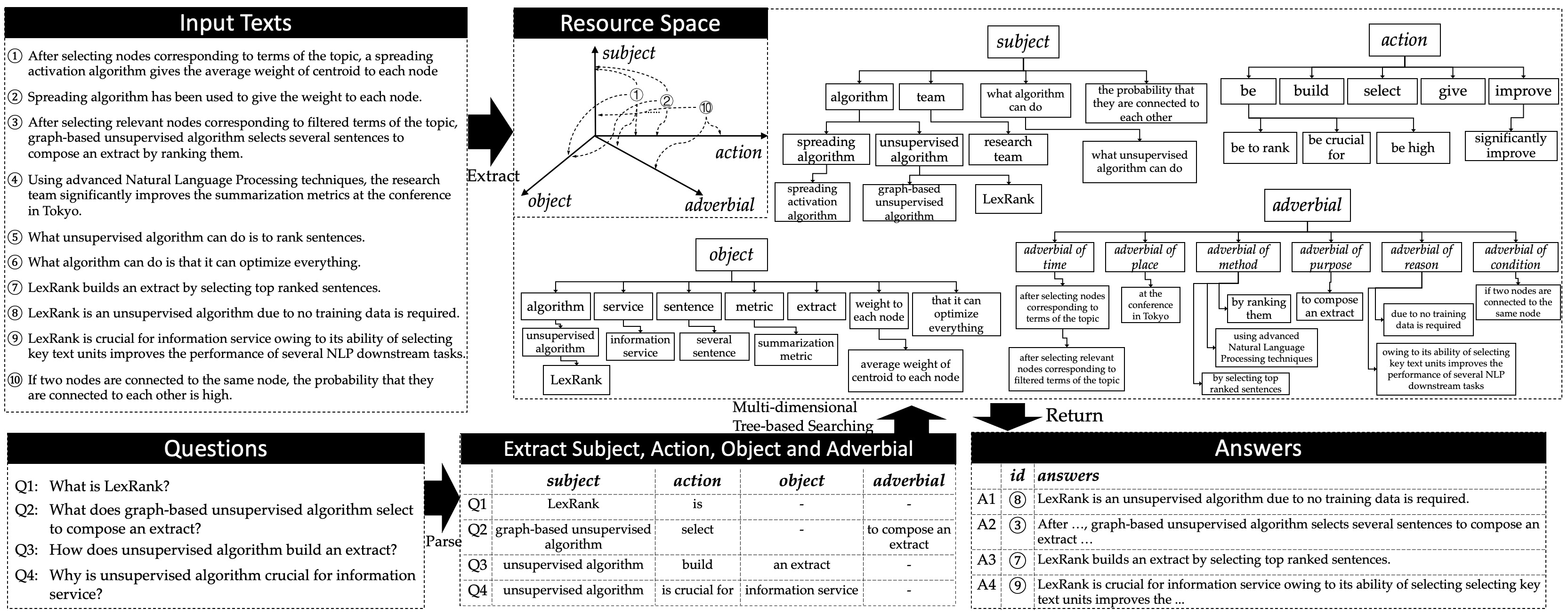}
  \caption{Examples of extracting dimensions according to syntax and query resource space.}
  \label{fig:fig2}
\end{figure}

\section{Extracting Subclass Relations}
\subsection{Identifying Patterns of Subject, action, object, and adverbial}
Identifying the patterns of subject, action, object, and adverbial is the basis for extracting subclass relations between subjects, between actions, between objects, and between adverbials.  A pattern is formed by abstracting on the structure of a representation and its syntax elements.  In English, the pattern of a sentence is formed by abstracting representation components into subjects, actions, objects and adverbials according to their roles and order in constructing sentence, forming the high-level pattern: <sentence>::= \{<adverbial>\} <subject> <action> [<object>] \{<adverbial>\}, where <subject> represents a thing that performs its <action>; <object> represents a thing that receives <action> represented by a verb or verb phrase linking subject and object; and, <adverbial> modifies <action> with representation of time, place, method, purpose, reason or condition.  

Verbs are classified into four classes: 1) modal verbs to indicate a modality; 2) auxiliary verbs to add functional or grammatical meaning to the clause; 3) main verb to express an action or event, which can be used as the head of a verb phrase; and, 4) linking verb to link the subject and the noun phrase/adjective phrase that describes the state of the subject, which can be used as the head of a linking verb phrase \cite{huddleston2021student}.

The role of action in representation is to link the subject and object so its core representation (called head) is the main verb or linking verb.  To represent a complex action, the head needs modal verb, auxiliary verb, adverb phrase, adjective phrase and prepositional phrase (called pre-head) to represent the modality, tense, voice, or mood of the head, or modify the head with time, place, frequency, manner, degree, certainty, purpose and so on.  The head also needs adverb phrase, adjective phrase, prepositional phrase and infinitive clause (called post-head) to further modify the head and describe the state of subject.  The role of action in representation determines its pattern as follows:

<action> ::= {<pre-head of action>} <head of action> {<post-head of action>}

<pre-head of action> ::= <modal verb>|<auxiliary verb>|<adverb phrase>|<adjective phrase>| <prepositional phrase>

<head of action> ::= <main verb> | <linking verb>

<post-head of action> ::= <adverb phrase>| <adjective phrase>|<prepositional phrase>|<infinitive clause>

<modal verb> ::= “can” |“could” |“may” |“might” |“must” |“should” |“shall” |“will” |“would” 

<auxiliary verb> ::= “have”| “has”| “had”| “am”| “is”| “are”| “was”| “were”| “be”| “been”| “being”

<adverb phrase> ::= <adverb>|<adverb>[<conjunction>]<adverb phrase>

<adjective phrase> ::= [<adverb phrase>] <adjective> [<prepositional phrase>] | <adjective phrase>[<conjunction>]<adjective phrase>

<infinitive clause> ::= “to”<verb phrase>

<prepositional phrase> ::= <preposition>[<verb phrase>]

<verb phrase> ::= [<adverb phrase>]<verb>[<adverb phrase>]

Both <pre-head of action> and <post-head of action> have a complex form with combination of their representations respectively, e.g., “the algorithm might have not been able to rank the sentences” where the underline part is the pre-head of the head “rank” and “syntax features have been applied primarily to identify relations” where the underline part is the post-head of the head “applied

Subject represents a person or a thing that controls the action, represented by nominal consisting of noun phrase and noun clause.  Noun phrase consists of noun, noun phrase, and pronoun. Noun clause consists of clause with subordinate conjunction, infinitive clause, and gerund clause.  A clause consists of lead word (i.e., subordinate conjunction), subject, action, object, and adverbial.  Therefore, the pattern of subjects takes the following form:

<subject> ::= <noun phrase> | <noun clause>

<noun phrase> ::= <noun phrase*> | <pronoun>

<noun clause> :: = [<subordinate conjunction>] [<subject>] <action> [<object>] {<adverbial>} | “to” <action> [<object>] {<adverbial>} | <gerund> [<object>] {<adverbial>}

<subordinate conjunction> ::= “that”| “whether”| “whom”| “whose”| “who”| “whoever”| “what”| “whatever”| “which”| “whichever”| “why”| “when”| “whenever”| “where”| “wherever”| “how”| “however”

The object of a sentence represents a person or thing to which the action of the sentence is directed.  It can be: 1) direct object, which receives action directly, represented by nominal; 2) direct object with indirect object, which receives action indirectly, represented by nominal (e.g., “the algorithm gives the weight to each node” where the underline part is the indirect object and the double underline part is the direct object); or, 3) direct object with object complement, which modifies direct object, represented by noun phrase or adjective phrase (e.g., “the algorithm makes results excellent” where the underline part is the object complement to describe the direct object “results”).  Therefore, the pattern of objects takes the following form:

<object> ::= <direct object> | <indirect object> <direct object> | <direct object> <preposition> <indirect object> | <direct object> <object complement>

<direct object> ::= <noun phrase>|<noun clause>

<indirect object> ::= <noun phrase>|<noun clause>

<object complement> ::= <noun phrase> | <adjective phrase>

<preposition> ::= “to” | “for” | “at” | “on”

The adverbial represents time, place, method, purpose, reason, and condition of an action represented by adverbial clause and adverbial phrase.  Its pattern is presented in Appendix A.

\subsection{Identifying Subclass Relations}
As representations of subject, action, object, and adverbial consist of phrases and clauses, the subclass relations between subjects, between actions, between objects and between adverbials are determined by the subclass relations between phrases and between clauses. The following rules identify the subclass relations between phrases, between clauses and between sentences respectively.

\textbf{Rule 1}. For $phrase_{1}$ represented as ($pre_{1}$, $head_{1}$, $post_{1}$) and $phrase_{2}$ represented as ($pre_{2}$, $head_{2}$, $post_{2}$), $phrase_{1}$ is the subclass of $phrase_{2}$ if 1) $head_{1}$ and $head_{2}$ are the same, and 2) $pre_{2} \cup poat_{2}$ is the subset of $pre_{1} \cup poat_{1}$.

For example, Rule 1 can identify that “graph-based unsupervised algorithm” is the subclass of “unsupervised algorithm”.

\textbf{Rule 2}. For verb phrases $vp_{1}$ represented as ($action_{1}$, $np_{1}$) and $vp_{2}$ represented as ($action_{2}$, $np_{2}$), $vp_{1}$ is the subclass of $vp_{2}$ if at least one of the pairs ($action_{1}$, $action_{2}$) and ($np_{1}$, $np_{2}$) satisfies the condition that the former is a subclass of the latter and the former and the latter of the other pair are the same.

For example, Rule 2 can identify that “automatically using graph-based unsupervised algorithm” is the subclass of “using graph-based algorithm”.

\textbf{Rule 3}. For prepositional phrases $prep_{1}$ represented as ($preposition_{1}$, $np_{1}$) and $prep_{2}$ represented as ($preposition_{2}$, $np_{2}$), $prep_{1}$ is the subclass of $prep_{2}$ if 1) $preposition_{1}$ and $preposition_{2}$ are the same, and 2) $np_{1}$ is a subclass of $np_{2}$.

For example, Rule 3 can identify that “through graph-based unsupervised algorithm” is the subclass of “through graph-based algorithm”.

\textbf{Rule 4}. For $clause_{1}$ represented as ($leadword_{1}$, $subject_{1}$, $action_{1}$, $object_{1}$, $adverbial_{1}$) and $clause_{2}$ represented as ($leadword_{2}$, $subject_{2}$, $action_{2}$, $object_{2}$, $adverbial_{2}$), $clause_{1}$ is the subclass of $clause_{2}$ if 1) $leadword_{1}$ and $leadword_{2}$ are the same, and 2) at least one of the following pairs ($subject_{1}$, $subject_{2}$), ($action_{1}$, $action_{2}$), ($object_{1}$, $object_{2}$) and ($adverbial_{1}$, $adverbial_{2}$) satisfies the condition that the former syntax element is a subclass of the latter syntax element and the former syntax element and the latter syntax element of each rest pair are the same.

For example, Rule 4 can identify that “what graph-based unsupervised algorithm can do” is the subclass of “what unsupervised algorithm can do”.

\textbf{Rule 5}. For $sentence_{1}$ represented as ($subject_{1}$, $action_{1}$, $object_{1}$, $adverbial_{1,1}$, …, $adverbial_{1,n}$) and $sentence_{2}$ represented as ($subject_{2}$, $action_{2}$, $object_{2}$, $adverbial_{2,1}$, …, $adverbial_{2,m}$), $sentence_{1}$ is the subclass of $sentence_{2}$ if 1) $n\geq m$, and 2) at least one of the following pairs ($subject_{1}$, $subject_{2}$), ($action_{1}$, $action_{2}$), ($object_{1}$, $object_{2}$), ($adverbial_{1,1}$, $adverbial_{2,1}$), …, and ($adverbial_{1,m}$, $adverbial_{2,m}$) satisfies the condition that the former syntax element is subclass of the latter syntax element and the former syntax element and the latter syntax element of each rest pair are the same. 

For example, Rule 5 can identify that “in China, researchers of ICT have published many papers about neural networks” is the subclass of “researchers have published many papers”.

\textbf{Rule 6}.  For $question_{1}$ represented as ($interrogative word_{1}$, $subject_{1}$, $action_{1}$, $object_{1}$, $adverbial_{1,1}$, …, $adverbial_{1,n}$) and $question_{2}$ represented as ($interrogative word_{2}$, $subject_{2}$, $action_{2}$, $object_{2}$, $adverbial_{2,1}$, …, $adverbial_{2,m}$), $question_{1}$ is the subclass of the $question_{2}$ if 1) $interrogative word_{1}$ is the same as $interrogative word_{2}$; 2) $n\geq m$; and, 3) at least one of the following pairs ($subject_{1}$, $subject_{2}$), ($action_{1}$, $action_{2}$), ($object_{1}$, $object_{2}$), ($adverbial_{1,1}$, $adverbial_{2,1}$), …, and ($adverbial_{1,m}$, $adverbial_{2,m}$) satisfies the condition that the former syntax element is subclass of the latter syntax element and the former syntax element and the latter syntax element of each rest pair are the same.

Based on the extracted subclass relations, subject abstraction tree, action abstraction tree, object abstraction tree and adverbial abstraction tree can be constructed by 1) combining the same language representation of subject, action, object, and adverbial, and 2) removing the subclass relations that can be derived from existing subclass relations to make simpler abstraction trees.

\subsection{Completeness}
The completeness of the patterns of representation is guaranteed by the completeness of the patterns of subject, action, object, and adverbial as the representation of a sentence consisting of subject, action, object, and adverbial.

The pattern of subject defines two types of nominals: 1) noun phrase consisting of noun/noun phrase and pronoun, and 2) noun clause consisting of relative clause, that-clause, infinitive clause, and gerund clause \cite{carter2006cambridge}, which cover all types of subjects.  All subjects can be extracted with the pattern from a given text following grammar.

The pattern of action defines a verb phrase that consists of 1) a head (a linking verb or a main verb); 2) pre-head consisting of modal verb (for representing modality and mood), auxiliary verb (for representing tense, voice and mood), adjective phrase (for representing modality and modifying head) and adverb phrase (for modifying head); and 3) post-head consisting of adjective phrase, adverb phrase, prepositional phrase and infinitive clause, which cover all types of actions.

The pattern of the object consists of three types: 1) direct object; 2) direct object with indirect object; and 3) direct object with object complement, which cover all types of objects in English. Direct object and indirect object are both nominals consisting of noun phrase and noun clause. The object complement modifies the direct object with noun phrase and adjective phrase \cite{carter2006cambridge}.

The pattern of adverbial defines two types of representations that cover all types of adverbials: 1) phrase (consisting of noun phrase, prepositional phrase, verb phrase, adverb phrase and phrase introduced with a subordinate conjunction), and 2) clause (consisting of clause introduced with a subordinate conjunction and infinitive clause).

These four patterns are not overlapped so that each pattern can be independently used.

\section{Patterns of Question and Patterns of Answer}
It is rational to assume that the questions and answers are all represented in sentences that follow English grammar.

\subsection{Patterns of Question}
In English grammar, the question consists of special question (i.e., wh-question) and general question (i.e., yes/no question).  The special question asks about specific things such as person, place, time, reason, purpose and method represented as subject, action, object, and adverbial with an interrogative word.  The general question starts with a modal verb, or an auxiliary verb to get a declarative sentence with affirmative or negative attitude.  As object can be direct object, indirect object and object complement, a question about object consists of question about direct object, question about indirect object, and question about object complement.   Thus, the high-level pattern of question in English takes the following form:

<question> ::= <special question> | <general question>

<special question> ::= <question about subject> | <question about object> | <question about adverbial>

<question about object> ::= <question about direct object> | <question about indirect object> | <question about object complement>

The <question about object> and <question about adverbial> (consisting of when, where, why, how) are presented in Appendix B.

The <question about subject> is used to know subject of a sentence (consisting of thing and person).  This pattern is formed by adding an interrogative word before the subject of sentence (declarative sentence). The following is the pattern of <question about subject>.

<question about subject> ::= \{<adverbialQ>\} <interrogative word> [<subjectQ>] <actionQ> [<objectQ>] \{<adverbialQ>\}

<interrogative word> ::= “what” | “who” | “which” | “whose”

<subjectQ> ::= <noun phrase>

<actionQ> ::= <action>

<objectQ> ::= <object>

<adverbialQ> ::= <adverbial>

An example of this pattern is “in search engine, what database stores the metadata?”, where “database” is <subjectQ>, “store” is <actionQ>, “metadata” is <objectQ> and “in search engine” is <adverbialQ>.

The <general question> is used to confirm or deny a declarative sentence. This pattern is formed by adding an auxiliary or a modal verb as an interrogative word before the subject of sentence.  The following is the pattern of <general question>.

<general question> ::= \{<adverbialQ>\} <interrogative word> <subjectQ> <action> [<objectQ>] \{<adverbialQ>\}

<interrogative word> ::= <auxiliary> | <modal>

<subjectQ> ::= <subject>

<actionQ> ::= [<interrogative word>] <action>

<objectQ> ::= <object>

<adverbialQ> ::= <adverbial>

An example is “In summarization, does graph-based algorithm incorporate structured representations?”, where “graph-based algorithm” is <subjectQ>, “incorporate” is <actionQ>, “structured representations” is <objectQ> and “in summarization” is <adverbialQ>

These patterns enable questions in natural language sentences to be understood and processed to select the correct answers.

\subsection{Properties of Answer}
A sentence is the answer to a question if it has the following properties:
\begin{enumerate}
\item \textit{Relevancy}.  The answer must contain at least one of the syntax elements (i.e., subject, action, object or adverbial) the question asked for such that the syntax elements of the answer are the same, synonym or subclass of the syntax elements of the question.

\item \textit{Consistency}. If the question is affirmative, it is better to have an affirmative answer. If the question is negative, it is preferable to have a negative answer. For example, for the question “what algorithm needs labelled data”, the sentences “unsupervised algorithm does not need the labelled data” is not a good answer as the question is affirmative while the answer is negative.  This characteristic can be used to select a good answer from a set of related answers.

\item \textit{Convertibility}. The voice of the answer can be the same as the question or in a different voice.  This characteristic can be used to judge relevant answers because the subject in one voice becomes object in another voice.   As all passive voices can be converted into active voices by pre-processing texts, this paper focuses on processing active voice only.
\end{enumerate}

The pattern of answer can be identified according to the above properties.

\subsection{Patterns of Answer}
The relevancy requests an answer to have the following commonalities in syntax elements of question about subject: the subject of answer is the subclass of the subject of its question, the action of answer is the same as, the synonym or subclass of the action of its question, the object of answer is the same as or subclass of the object of its question, and the adverbial of answer is the same as or subclass of the adverbial of its question.  Therefore, the pattern of the answer to <question about subject> is formed as follows.

<answer of question about subject> ::= {<adverbialA>} <subjectA> <actionA> [<objectA>] {<adverbialA>}

<subjectA> ::= <subclass of <subjectQ>>

<actionA> ::= <actionQ> | <synonym of <actionQ>> | <subclass of <actionQ>> | <subclass of <actionA>>

<objectA> ::= <objectQ> | <subclass of <objectQ>>

<adverbialA> ::= <adverbialQ> | <subclass of <adverbialQ>>

An example of this pattern is “database in master service stores metadata of the web pages in master-slave search engine” (it answers “In search engine, what database stores the metadata?”), where “database in master service” is <subjectA> which is the subclass of <subjectQ>, “store” is <actionA> which is the same as <actionQ>, “metadata of the web pages” is <objectA> which is the subclass of <objectQ>, “in master-slave search engine” is <adverbialA> which is the subclass of <adverbialQ>.

The pattern of the answer to the question about object and the pattern of the answer to the question about adverbial are presented in Appendix B.

The pattern of answer to the general question is the same as the pattern of sentence that satisfies: (1) the subject of answer is the same as or the subclass of the subject of its question, (2) the action of answer is the same as, the synonym or subclass of the action of its question, (3) the object of answer is the same as or subclass of the object of its question, and (4) the adverbial of answer is the same as or subclass of the adverbial of its question.

\subsection{Completeness of the Patterns of Question and Answer}
The pattern of question consists of 1) special question, and 2) general question, which cover all types of question in English.  The purpose of special question is to know 1) thing or person (“what”, “which”, “who”, “whom” and “whose”), 2) time (“when”), 3) place (“where”), 4) reason (“why”), and 5) method (“how”), which cover all cases of asking specific things.  The general question is to get an affirmative or a negative judgement on the question, which covers all cases of answers.

The completeness of the pattern of the answer is also guaranteed by the three properties of answer.  If a sentence does not satisfy the relevancy, the semantics represented by the syntax elements of the sentence (action, the thing that performs the action, the thing that receives the action, time, place, method, purpose, reason, and condition of the action) are all different from the semantics represented by the syntax elements of the question, therefore it is not an answer to the question.  If a sentence does not satisfy the consistency, it does not answer the question directly, therefore it is not an answer to the question. 

So, the patterns of question and answer consider all cases.

\section{Relations between Questions and Answers}
For a set of questions, relations between questions restrict the relation between their answers because answers are closely related to their questions.  Knowing this relations helps determine the range of answers.

\textbf{Theorem 1}.  If $question_{1}$ is the subclass of $question_{2}$, the answer set of $question_{1}$ is the subset of the answer set of $question_{2}$.

\textit{Proof}.   If $question_{1}$ represented as ($interrogative word_{q1}$, $subject_{q1}$, $action_{q1}$, $object_{q1}$, $adverbial_{q1,1}$, …, $adverbial_{q1,n}$) is the subclass of $question_{2}$ represented as ($interrogative word_{q2}$, $subject_{q2}$, $action_{q2}$, $object_{q2}$, $adverbial_{q2,1}$, …, $adverbial_{q2,m}$), then for each of the following pairs ($subject_{q1}$, $subject_{q2}$), ($action_{q1}$, $action_{q2}$), ($object_{q1}$, $object_{q2}$), ($adverbial_{q1,1}$, $adverbial_{q2,1}$), …, and ($adverbial_{q1,m}$, $adverbial_{q2,m}$), the former syntax element is the subclass of or the same as the latter syntax element according to Rule 6, that is, the former syntax element is a subtree of or the same as the latter syntax element.  If all answer sentences are organized in an answer space $A(subject,action,object,adverbial)$, each dimension represents a set of answers denoted as $A(subject)$, $A(action)$, $A(object)$, and $A(adverbial)$.   A subclass relation between two syntax elements represents a subtree relation on the same dimension.  A subtree represents a subset of the answer set represented by its superclass on the same dimension, i.e., $A(X) \subseteq A(Y)$ if $X$ is a subclass of $Y$. The answer set determined by $question_{1}$ is $A(subject_{q1}) \cap A(action_{q1}) \cap A(object_{q1}) \cap A(adverbial_{q1,1}) \cap ... \cap A(adverbial_{q1,m})$, and answer set determined by $question_{2}$ is $A(subject_{q2}) \cap A(action_{q2}) \cap A(object_{q2}) \cap A(adverbial_{q2,1})\cap ... \cap A(adverbial_{q2,m})$, and $A(subject_{q1}) \subseteq A(subject_{q2})$, $A(action_{q1}) \subseteq A(action_{q2})$, $A(object_{q1}) \subseteq A(object_{q2})$, $A(adverbial_{q1,1}) \subseteq A(adverbial_{q2,1})$, ..., $A(adverbial_{q1,m}) \subseteq A(adverbial_{q2,m})$, so the answer set of $question_{1}$ is a subset of the answer set of $question_{2}$. $\blacksquare$

Generally, the subclass relation can be relaxed to relevant relation.

\textbf{Definition 1}.  A $question_{1}$ represented as ($interrogative word_{q1}$, $subject_{q1}$, $action_{q1}$, $object_{q1}$, $adverbial_{q1,1}$, …, $adverbial_{q1,n}$) is relevant (or irrelevant) to $question_{2}$ represented as ($interrogative word_{q2}$, $subject_{q2}$, $action_{q2}$, $object_{q2}$, $adverbial_{q2,1}$, …, $adverbial_{q2,m}$) if there exists (or does not exist) one pair ($n_{q1}$, $n_{q1}$) $\in$ \{($subject_{q1}$, $subject_{q2}$), ($action_{q1}$, $action_{q2}$), ($object_{q1}$, $object_{q2}$), ($adverbial_{q1,1}$, $adverbial_{q2,1}$), …, and ($adverbial_{q1,m}$, $adverbial_{q2,m}$)\} such that $n_{q1}$ is the subclass/superclass of or the same as $n_{q2}$.

\textbf{Definition 2}.  A $sentence_{1}$ represented as ($subject_{1}$, $action_{1}$, $object_{1}$, $adverbial_{1,1}$, …, $adverbial_{1,n}$) is relevant (or irrelevant) to $sentence_{2}$ represented as ($subject_{2}$, $action_{2}$, $object_{2}$, $adverbial_{2,1}$, …, $adverbial_{2,m}$) if there exists (or does not exist) one pair ($n_{1}$, $n_{1}$) $\in$ \{($subject_{1}$, $subject_{2}$), ($action_{1}$, $action_{2}$), ($object_{1}$, $object_{2}$), ($adverbial_{1,1}$, $adverbial_{2,1}$), …, and ($adverbial_{1,m}$, $adverbial_{2,m}$)\} such that $n_{1}$ is the subclass/superclass of or the same as $n_{2}$.

\textbf{Theorem 2}.  If $question_{1}$ is relevant to $question_{2}$, the answer of $question_{1}$ is relevant to the answer of $question_{2}$.

\textit{Proof}.   Let $question_{1}$ represented as ($interrogative word_{q1}$, $subject_{q1}$, $action_{q1}$, $object_{q1}$, $adverbial_{q1,1}$, …, $adverbial_{q1,n}$) and $question_{2}$ represented as ($interrogative word_{q2}$, $subject_{q2}$, $action_{q2}$, $object_{q2}$, $adverbial_{q2,1}$, …, $adverbial_{q2,m}$).  If $question_{1}$ is relevant to $question_{2}$, then there exists at least one syntax element (denoted as $X$) of $question_{1}$ which is the same as or a subclass of the same syntax element (denoted as $Y$) of $question_{2}$, that is, one syntax element is a subtree of or the same as the other.  If all answer sentences are organized in an answer space $A(subject,action,object,adverbial)$, each dimension represents a set of answer sentences denoted as $A(subject)$, $A(action)$, $A(object)$, and $A(adverbial)$. The answer sentence of $question_{1}$ should be represented by $X$ and the answer sentence of $question_{2}$ should be represented by $Y$ on the same dimension. According to Definition 2, the answer of $question_{1}$ is relevant to the answer of $question_{2}$. $\blacksquare$

According to Rule 5 and Theorem 1, we have the following corollary.

\textbf{Corollary 1}. If a sentence $s$ is the subclass of another sentence $s’$, and $s’$ is an answer of a question, then $s$ is also the answer sentence of the question.

The following corollary can be drawn from Definition 2 and Theorem 2, and the fact that a syntax element of two answer sentences may not appear in their questions.

\textbf{Corollary 2}.  If $question_{1}$ is irrelevant to $question_{2}$, the answer sentence of $question_{1}$ is irrelevant or relevant to the answer sentence of $question_{2}$.

\section{Verification of Extracted Class Trees}
To verify the correctness of the extracted abstraction trees and the feasibility of the abstraction trees in question answering application, an experiment is carried out on different types of texts.

\subsection{Datasets}
The experiment consists of the following six datasets:

\begin{enumerate}
    \item Summ, which consists of 100 scientific papers on summarization.
    \item AI, which consists of 372 scientific papers from the Artificial Intelligence journal (https://www.sciencedirect.com/journal/artificial-intelligence).
    \item ACL, which consists of 173 scientific papers (including 88 long papers and 85 short papers) from the Proceedings of the 52nd Annual Meeting of the Association for Computational Linguistics (https://aclanthology.org/events/acl-2014/).
    \item MED, which consists of 33 scientific papers (published in 2020) from the British Medical Journal (https://www.bmj.com/).
    \item DUC, which consists of 567 English news articles collected from TREC-9 with 59 topics in 2002 (https://www-nlpir.nist.gov/projects/duc/guidelines/2002.html).
    \item LEGAL \cite{galgani2012citation}, which consists of 3890 Australian legal case reports (from 2006 to 2009) from the Federal Court of Australia (https://archive.ics.uci.edu/ml/datasets/Legal+Case+Reports).
\end{enumerate}

\subsection{Verification of the Extracted Trees}
The subclass relations of extracted subject, action, object, and adverbial are evaluated by the standard metrics of relation extraction (consisting of precision, recall and F1-score, precision = correctly identified subclass relations / all subclass relations extracted by Rule 1-6, recall = correctly identified subclass relations / all the labelled subclass relations that should be identified, and F1-score is the harmonic mean of precision and recall). As shown in TABLE 2-5, the average precision, recall and F1-score of the subclass relations of the extracted actions and subjects of six datasets are all greater than 90.00\%, the average precision, recall and F1-score of the subclass relations of the extracted objects and adverbials of six datasets are all greater than 80.00\%.

\begin{table}
 \caption{Experimental results of the extracted subclass relations of actions.}
  \centering
  \begin{tabular}{llll}
    \toprule
    \textbf{Dataset}  &	\textbf{Precision}    &   \textbf{Recall}  &	\textbf{F1-score} \\
    \midrule
    Summ    &   96.07\%	& 92.45\%	& 94.23\% \\
    AI	    &   91.78\%	& 89.93\%	& 90.85\% \\
    ACL	    &   93.33\%	& 89.36\%	& 91.30\% \\
    MED	    &   94.23\%	& 90.74\%	& 92.45\% \\
    DUC	    &   92.59\%	& 90.91\%	& 91.74\% \\
    LEGAL   &	93.02\%	& 91.39\%	& 92.20\% \\
    \midrule
    Avg.    &   93.50\%	& 90.80\%	& 92.13\% \\
    \bottomrule
  \end{tabular}
  \label{tab:table2}
\end{table}

\begin{table}
 \caption{Experimental results of the extracted subclass relations of subjects.}
  \centering
  \begin{tabular}{llll}
    \toprule
    \textbf{Dataset}  &	\textbf{Precision}    &   \textbf{Recall}  &	\textbf{F1-score} \\
    \midrule
    Summ	&94.11\%	&90.57\%	&92.31\%  \\
    AI	&93.15\%	&91.28\%	&92.20\%  \\
    ACL	&92.22\%	&88.30\%	&90.22\%  \\
    MED	&95.19\%	&91.67\%	&93.40\%  \\
    DUC	&92.06\%	&90.63\%	&91.34\%  \\
    LEGAL	&92.90\%	&89.95\%	&91.40\%  \\
    \midrule
    Avg.	&93.27\%	&90.40\%	&91.81\%  \\
    \bottomrule
  \end{tabular}
  \label{tab:table3}
\end{table}

\begin{table}
 \caption{Experimental result of the extracted subclass relations of objects.}
  \centering
  \begin{tabular}{llll}
    \toprule
    \textbf{Dataset}  &	\textbf{Precision}    &   \textbf{Recall}  &	\textbf{F1-score} \\
    \midrule
    Summ	&87.25\%	&83.96\%	&85.58\%  \\
    AI	&81.51\%	&79.87\%	&80.68\%  \\
    ACL	&84.44\%	&80.85\%	&82.61\%  \\
    MED	&87.50\%	&84.26\%	&85.85\%  \\
    DUC	&84.62\%	&82.09\%	&83.33\%  \\
    LEGAL	&81.48\%	&79.57\%	&80.51\%  \\
    \midrule
    Avg.	&84.47\%	&81.77\%	&83.09\%  \\
    \bottomrule
  \end{tabular}
  \label{tab:table4}
\end{table}

\begin{table}
 \caption{Experimental result of the extracted subclass relations of adverbial.}
  \centering
  \begin{tabular}{llll}
    \toprule
    \textbf{Dataset}  &	\textbf{Precision}    &   \textbf{Recall}  &	\textbf{F1-score} \\
    \midrule
    Summ	&84.62\%	&81.48\%	&83.02\%  \\
    AI	&86.21\%	&80.65\%	&83.33\%  \\
    ACL	&83.33\%	&81.08\%	&82.19\%  \\
    MED	&85.71\%	&82.76\%	&84.21\%  \\
    DUC	&84.06\%	&80.33\%	&82.16\%  \\
    LEGAL	&83.06\%	&79.45\%	&81.22\%  \\
    \midrule
    Avg.	&84.50\%	&80.96\%	&82.69\%  \\
    \bottomrule
  \end{tabular}
  \label{tab:table5}
\end{table}

\subsection{Coverage of Extracted Class Trees}
The coverage of the four extracted class trees (the number of sentences covered by the abstraction tree/all sentences where the tree is extracted) is shown in TABLE 6, where the average coverage of the action is 96.53\%, the average coverage of the subject is 91.19\% (where there are several imperative sentences), the average coverage of the object is 86.75\% (where there are several sentences without objects), and the average coverage of the adverbial is 63.56\% (where there are some sentences without adverbials).  The coverage of the sentences represented by subject, action, object, and adverbial is 100\%.  The average coverage of the sentences commonly represented by subject, action and object is 85.75\%.  The high coverage of the four abstraction trees is the basis for supporting question answering application.

\begin{table}
 \caption{Coverage of three class trees.}
  \centering
  \begin{tabular}{llllll}
    \toprule
    \textbf{Dataset}	&\textbf{Action}	&\textbf{Subject}	&\textbf{Object}	&\textbf{Adverbial}	&\textbf{Coverage of Trees}  \\ \\
    \midrule
    Summ	&96.87\%	&92.77\%	&87.71\%	&64.52\%	&100.00\%  \\
    AI	&97.20\%	&89.44\%	&89.13\%	&63.77\%	&100.00\%  \\
    ACL	&96.19\%	&90.29\%	&88.57\%	&61.84\%	&100.00\%  \\
    MED	&94.20\%	&90.46\%	&84.11\%	&60.31\%	&100.00\%  \\
    DUC	&96.77\%	&90.32\%	&82.26\%	&66.79\%	&100.00\%  \\
    LEGAL	&97.95\%	&93.85\%	&88.72\%	&64.15\%	&100.00\%  \\
    \midrule
    Avg.	&96.53\%	&91.19\%	&86.75\%	&63.56\%	&100.00\%  \\
    \bottomrule
  \end{tabular}
  \label{tab:table6}
\end{table}

\section{Verification of Effect of Question Answering}
Verification is conducted on the Summ dataset consisting of 5670 words and 306 sentences on average.  Figure 3 shows the process of extracting dimensions from the dataset and querying from the dimensions. The rectangles in white and green represent noun phrases, and the rectangles in grey represent clauses. The subclass relations between white and green rectangle are identified by syntactic pattern-based approach, and the rest subclass relations are identified by Rule 1-6.  The subject dimension has 12893 nodes and 7172 subclass relations.  The action dimension has 7975 nodes and 6379 subclass relations.  The object dimension has 18304 nodes and 8388 subclass relations.  The adverbial dimension has 10555 nodes and 7527 subclass relations.

\begin{figure}
  \centering
  \includegraphics[width=1\linewidth]{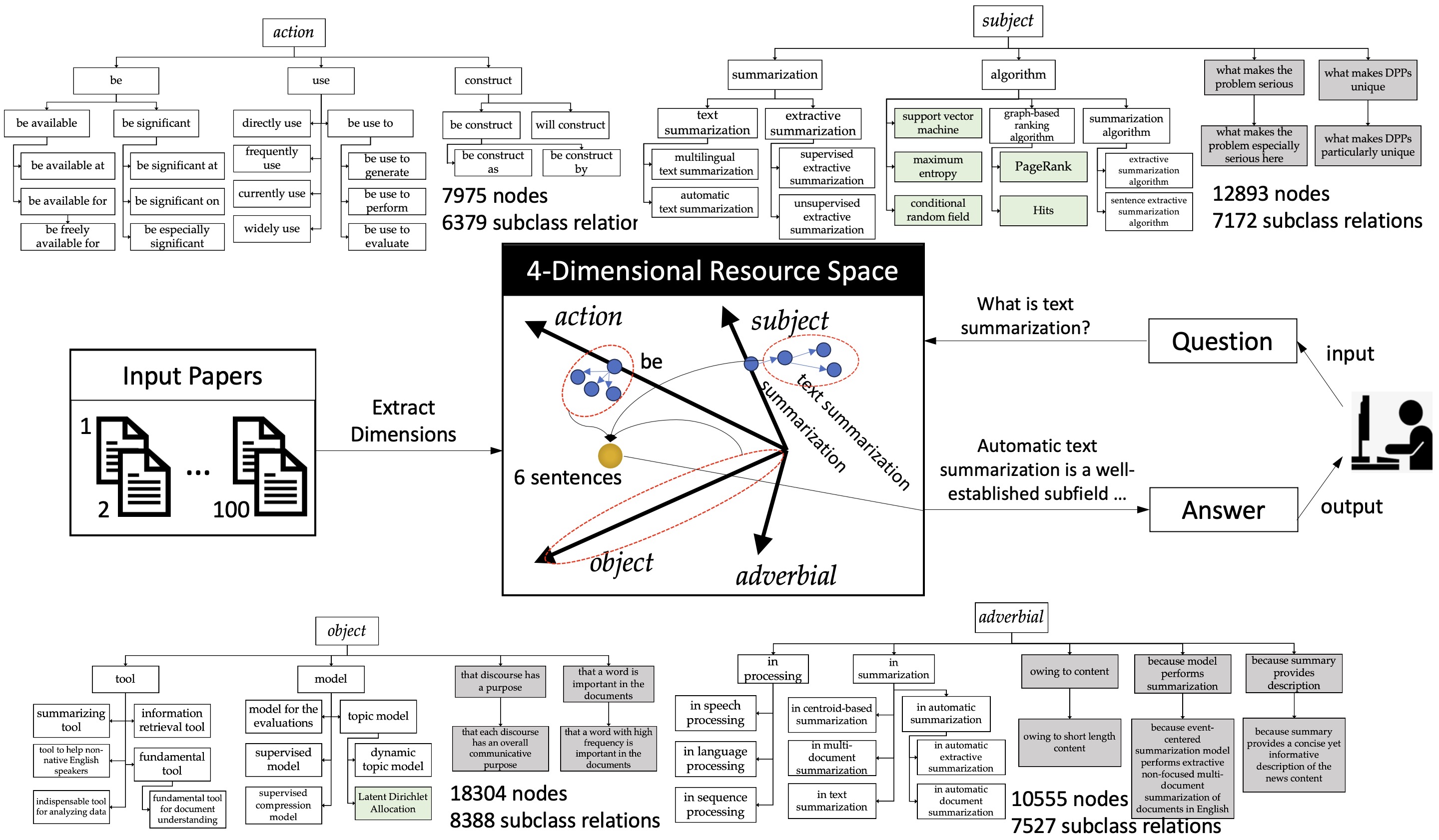}
  \caption{General process of extracting four dimensions from Summ dataset and process of querying the space with the four dimensions in natural language.}
  \label{fig:fig3}
\end{figure}

For the question “What is text summarization?”, six answer sentences are located according to the pattern of the answer of the what question, as shown in TABLE 7, by 1) searching from subject dimension with subject “text summarization”; 2) searching from action dimension with action “be”; and, 3) searching from object dimension with its root, i.e., object (object should not be empty as it is the question asked for).  The precision of the six answer sentences is 100\%.

For the question “What does general corpus represent in NLP task?”, one correct answer sentence is located as shown in TABLE 7.

\begin{table}
  \caption{Examples of answering questions.}
  \centering
  \begin{tabular}{>{\raggedright\arraybackslash}p{2cm} >{\raggedright\arraybackslash}p{3cm} >{\raggedright\arraybackslash}p{8cm} }
    \toprule
    \textbf{ID} &	\textbf{Question} &	\textbf{Answer} \\
    \midrule
    Q1 & What is text summarization? & 1. Automatic text summarization is a well-established subfield of natural language processing, which is relevant for a number of scenarios. \newline
2. Automatic text summarization is an essential tool in this era of information overloading. \newline
3. Automatic text summarization is the process of abstracting large texts into a few paragraphs while preserving its information content. \newline
4. Automatic text summarization is a seminal problem in information retrieval and natural language processing. \newline
5. Automatic text summarization is one of the widely used applications in the field of natural language processing (NLP). \newline
6. Text summarization is the process of selecting the most salient information in one or more textual documents. \\
    \midrule
    Q2 & What does general corpus represent in NLP task? &	In NLP tasks, a language is represented by a huge general corpus in that language.\\
    \bottomrule
  \end{tabular}
  \label{tab:table7}
\end{table}

We designed 300 questions of different types and 366 answers are annotated (the question with the most answers has 12 answers and the question with the least answers has one answer).  TABLE 8 shows the precision compared with the common-word-based approach and the BM25 approach (the top 5 answers are selected).  The result shows that the proposed approach achieves better performance than the other two approaches.  The common-word-based approach measures the relevance of a sentence and a question based on the number of common words while the BM25 approach measures the relevance of a sentence and a question by term frequency of the common words, inverse document frequency of the common words, and the length of the sentence.  The high relevance between a sentence and a question measured by these approaches is not a sufficient condition for the sentence to be the answer to the question.  The proposed approach is based on linguistic characteristics and selects the answer sentences based on the relations between subject, action, object, and adverbial within the texts and question.

\begin{table}
  \caption{Precision of different approaches on question answering.}
  \centering
  \begin{tabular}{ll}
    \toprule
    \textbf{Approach} &	\textbf{Precision} \\
    \midrule
    Common-word-based &	78.41\% \\
    BM25 &	83.06\%  \\
    The Proposed Approach &	92.08\% \\
    \bottomrule
  \end{tabular}
  \label{tab:table8}
\end{table}

\section{Related Work and Discussion}
\subsection{Subclass Relation Extraction}
Previous approaches to extracting subclass relations consist of syntactic-pattern approach, word-embedding approach, deep-learning-approach and large-language-model approach.

The syntactic-pattern approach summarizes empirical syntactic patterns that indicate subclass relations. Different types of patterns were developed according to different representations, e.g., for noun phrase, “$NP_{hyper}$ such as $NP_{hypon}$” \cite{hearst1992automatic}, “$NP_{hypon}$ is a $NP_{hyper}$”, “$NP_{hypon}$ and other $NP_{hyper}$”\cite{etzioni2004web}, and “$NP_{hyper}$ are $NP_{hypon}$ that” \cite{kozareva2010semi}; and, for verb phrase, “if he is $VP_{hypon}$-ing, then he is $VP_{hyper}$-ing” , “$VP_{hypon}$-ing is a kind of $VP_{hyper}$-ing” and “to $VP_{hypon}$ is to $VP_{hyper}$” \cite{huminski2018wordnet,lo2008automatic}.  The approach has the advantage of summarizing the knowledge of representing subclass relations but 1) it can only extract subclass relations within the same sentence, 2) the quality of the extraction is low (the average maximum F-scores of the Hearst Patterns is 15\% \cite{snow2004learning}), and 3) syntactic patterns are unsuitable for scientific literature, news articles and legal case reports, especially syntactic patterns for verb phrases, which almost do not exist within these texts.

The word-embedding approach is based on the assumption that subclass relations can be represented by embedding vectors, such as Word2Vec, Glove, or BERT embeddings \cite{devlin2019bert, mikolov2013distributed, pennington2014glove}.  This approach transforms the problem of subclass relation extraction into the problem of classification, utilizing combinations of word embedding vectors (such as concatenation\cite{baroni2012entailment}, difference, addition, and element-wise product \cite{weeds2014learning}) as features to train a classifier model (e.g., Support Vector Model) to predict subclass relations.  However, a work demonstrated that subclass relations of verbs are not identified by the logistic regression classifier trained based on combination (including difference and element-wise product) of word embedding vectors extracted from BERT \cite{narkevich2021bert}.  It is difficult for the word embedding based approach to directly extract subclass relations from texts, it requires word pairs that may have potential subclass relations in advance.  It cannot distinguish a word with different meanings by Word2Vec or Glove embeddings. It cannot process phrases or clauses and lacks interpretability. In contrast, the proposed approach is based on the pattern of phrases, clauses, and the abstraction relations between phrases and between clauses.

The deep learning approach that extracts subclass relations from texts involve two stages: named entity recognition and relation extraction.  The approach trains a named entity recognition model with sequence labeling texts to extract words or phrases whose subclass relation is to be determined further, and it regards the relation extraction problem as a classification problem. The approach predicts the subclass relations of two words or phrases by annotating examples to train a classification model. Commonly used named entity recognition and relation extraction models are based on RNN \cite{zhou2016attention}, CNN \cite{zeng2014relation} and transformer structured model such as BERT \cite{vaswani2017attention}. These approaches depend on a large number of training datasets and lack interpretability and controllability. The scope of extracting subclass relation is limited by the maximum input length accepted by the model. Therefore, it is difficult for deep learning approaches to extract the subclass between subjects, actions, objects and adverbials in different sentences.  In contrast, the proposed approach has a good interpretability and is unsupervised as it is based on linguistic characteristics (i.e., syntax pattern of the subject, action, object, and adverbial), and the scope of extracting the subclass relations is not limited to a small scale of representation such as a sentence.

The LLM-based approach predicts the sequence of words that describe relations based on prompt and input texts.  The following is input (consisting of an example [Example Texts] [Example Subclass Relations], instruction [Task] and input text [Input Texts]) and output [Output] of LLM to extract the subclass relations.

Example Texts: Healthcare question answering (HQA) is a challenging task as questions are generally non-factoid in nature.

Example Subclass Relations: (healthcare question answering, question answering), (challenging task, task)

Task: extract subclass relations of the following texts.

Input Texts: Automatic summarization of texts is now crucial for several information retrieval tasks owing to the huge amount of information available in digital media, which has increased the demand for simple, language-independent extractive summarization strategies.

Output: Subclass Relations: 

(Automatic summarization, text summarization)

(Information retrieval tasks, tasks)

(Language-independent extractive summarization strategies, extractive summarization strategies)

However, the output subclass relation pair (automatic summarization, text summarization) is incorrect, and the pairs (digital media, media) and (simple, language-independent extractive summarization strategies, strategies) are lost.  The performance of LLM-based systems depends on the prompt, without understanding subclass relation.  It predicts the next words based on the updated weights influenced by the prompt, which may generate incorrect answers without satisfying the request.  LLM-based systems are not specialized in identifying subclass relations; therefore, the performance (precision and recall) of the extracted subclass relations may be insufficient. It lacks interpretability, e.g., it cannot explain why the subclass relation pair (automatic summarization, text summarization).  Although the input length of the latest LLM-based systems has been significantly extended, it is still difficult for them to extract subclass relations for large-scale input texts, e.g., thousands of scientific papers.

The proposed approach is faithful to the input texts as it is based on linguistic characteristics and answers questions based on the relations between subject, action, object, and adverbial in input texts.

\subsection{Open Information Extraction}
Unsupervised approaches to open information extraction consist of rule-based approach and clause-based approach.

The rule-based approach uses rules to extract relation tuples. For example, an approach uses part-of-speech to define the relation in form of the following types: 1) a single verb, 2) a verb with a preposition, or 3) a verb followed by nouns/adjectives/adverbs and ending with a preposition. The nearest noun phrases on the left and right sides of the relation are extracted as the entities to compose a relation triple \cite{fader2011identifying}.

The clause-based approach extracts relation tuples by 1) extracting the set of independent clauses of the sentence, and 2) transforming each clause into a relation tuple according to clause pattern. For example, an approach (ClauseIE) extracts the independent clauses through the dependency parse tree by mapping the dependency relations into clause constituents, e.g., mapping dependency relation “nsubj” into the subject of clause, and mapping dependency relations “dobj”, “iobj”, “xcomp”, or “ccomp” into the object of clause. It returns relation tuple according to 7 clause types, e.g., the clause type “SV” represents a subject with a verb and return a 2-tuple (the clause “Mike died” returns (“Mike”, “died”)), the clause type “SVOO” represents a subject with a verb, a direct object and an indirect object and return a 4-tuple (the clause “The school awarded Mike a scholarship.” returns (“The school”, “awarded”, “Mike”, “a scholarship”)) \cite{colliat1996olap}.  The Stanford OpenIE can extract entities and their relation from sentences in form of triples based on the dependency parse tree \cite{angeli2015leveraging}. It traverses the dependency parse tree to split each sentence into a set of independent clauses through a classifier by predicting each edge whether it can yield an independent clause or not, and then each independent clause is transformed into a triple based on hand-crafted patterns, e.g., the pattern that the input “there are <$entity_{1}$> with <$entity_{2}$>” is transformed into the triple (<$entity_{1}$>, “have”, <$entity_{2}$>). However, it does not focus on the syntax pattern of the sentence (consisting of subject, action, object) and relations between entities from different independent clauses. The triples ignore adverbials that modify action.  The clause-based approaches such as Sandford OpenIE and ClauseIE fail to extract relation from the sentence “to solve complex problems is the essence of mathematics” where subject is a complex noun clause.  Examples comparing the extracted relations from different OpenIE systems with the proposed approach are presented in Appendix C.

\subsection{Automatic Construction of Resource Space and Normal Forms}
Automatically constructing resource space is a fundamental problem of developing Resource Space Model (RSM) \cite{zhuge2016multi}.  Eight criteria for discovering dimensions from texts have been introduced \cite{zhuge2016multi}.  One of them is the In-All criterion: If a representation can represent a set of texts, the representation or its specialization should appear within all texts of a class.

An approach uses candidate concepts extracted based on LDA, Wikipedia entities and WordNet from texts to construct Wikipedia hierarchy to construct dimensions \cite{yu2014framework}, however, it cannot handle phrases/clauses and relations that are not given in Wikipedia hierarchy and normal forms are not addressed.

To manage resources accurately and rigorously, normal forms of resource space are defined as the criteria for designing proper resource spaces \cite{zhuge2008web}.  The first-normal-form resource space (1NF) is that there does not exist duplicate name (in semantics) between coordinates at any dimension of the resource space. A coordinate of a dimension is a class identified by a node (e.g., noun phrase, noun clause and so on) or a class tree, where each class is identified by a path of classes started from its root. The proposed approach to extracting class trees from texts ensures that the same nodes are combined, which ensure that there does not exist the same path within any tree. Hence, the automatically constructed resource space is the 1NF resource space.  The second-normal-form resource space (2NF) is a 1NF resource space and the intersection of any two coordinates on the same dimension is empty for all dimensions of the resource space. According to the high-level pattern of the sentence, a sentence can have at most one subject, one action, and one object, which ensures that the subspace composed of subject tree, action tree and object tree is the 2NF resource space.  The third-normal-form resource space (3NF) is a 2NF resource space and all resources in the space can be managed by every dimension. The subject tree, action tree and object tree commonly cover more than 85\% of the texts and the rest less than 15\% of the texts can be covered by at least one of the dimensions. Hence, a subspace of the constructed resource space is a 3NF resource space.

Different from previous work on constructing resource space \cite{yu2014framework}, the proposed approach does not require external conditions such as the Wordnet, and does not need to assign the names of dimensions in advance.  Different from previous text clustering approaches such as those based on the Vector Space Model \cite{jain1988algorithms, salton1988term}, the proposed approach makes use of linguistic characteristics so the extracted abstraction trees can represent multiple abstractions of syntax, which enables the multi-dimensional resource space to efficiently access resources by subclass relations from different dimensions.  Previous work on multi-dimensional database and the dataspace do not concern automatic construction of the space nor linguistic characteristics \cite{colliat1996olap, franklin2005databases}.

\section{Conclusion}
This paper proposes an approach to automatically discover abstraction dimensions from texts according to syntax pattern. Each dimension is a class tree extracted by identifying the patterns of syntax elements (i.e., subject, action, object, and adverbial) and subclass relations judged by the patterns of their representations (i.e., phrases and clauses) and six rules of representing subclass relations.

The generated multi-dimensional resource space has a high coverage of the input texts; therefore, it can effectively operate texts from multiple dimensions. It supports question answering better than ranking-based approaches and generative approaches.  This paper shows that extracting abstraction dimensions from texts is significant and feasible.  The proposed approach does not rely on external condition but can be easily integrated with external conditions.

\appendix
\section{Patterns of Adverbial}
The adverbial is used to modify an action for additional representation of when, where, why, and how an action happens, consisting of time, place, method, purpose, reason, and condition. It is represented by adverbial clause and adverbial phrase.  So, the high-level structure of the pattern of adverbial is as follows:

<adverbial> ::= <adverbial of time> | <adverbial of place> | <adverbial of method> | <adverbial of purpose> | <adverbial of reason> | <adverbial of condition>

The following is the pattern of adverbial of time:

<adverbial of time> ::= <time phrase>| <time clause>

<time phrase> ::= <time noun phrase>| <time prepositional phrase>

<time noun phrase> ::= {<pre-modifier>} <time noun> {<post-modifier>}

<time prepositional phrase> ::= <prepositional phrase> <time noun phrase> 

<prepositional phrase> ::= <preposition>| <preposition> <adverb> <preposition>

<time clause> ::= <time subordinate conjunction> [<subject>] [<action>] [<object>] {<adverbial>}

<time subordinate conjunction> ::= “after”| “before”| “since”| “when”| “while”| “once”| “until”| “whenever”

<time noun>, <time phrase> and <time clause> represent the noun, phrase and clause that reflect time or date respectively, e.g., “week” (noun), “next week” (noun phrase), “in January” (prepositional phrase) and “when the optimization is done” (clause).

The following is the pattern of adverbial of place:

<adverbial of place> ::= <place phrase>| <place clause>

<place phrase> ::= <prepositional phrase> <place noun phrase>

<place noun phrase> ::= {<pre-modifier>} <place noun> {<post-modifier>}

<place clause> ::= <place subordinate conjunction> [<subject>] [<action>] [<object>] {<adverbial>}

<place subordinate conjunction> ::= “where”| “wherever”

<place noun>, <place phrase> and <place clause> represent the noun, phrase and clause that reflect place or location respectively, e.g., “China” (noun), “in Beijing” (prepositional phrase), “where the events happen” (clause).

The following is the pattern of adverbial of method:

<adverbial of method> ::= <method phrase>| <method clause>

<method phrase> ::= <method verb phrase> | <adverb phrase>

<method verb phrase> ::= <method action> <noun phrase>

<method action> ::= {<pre-head>} <method verb> {<post-head>}

<method verb> ::= “using”| “utilizing”| “employing”| “applying”| “exploiting”| “implementing”

<method clause> ::= <method subordinate conjunction> [<subject>] [<action>] [<object>] {<adverbial>}

<method subordinate conjunction> ::= “as”| “as if” | “as though”| “like” | “by”| “through”| “via”| “with”| “on”

<method phrase> and <method clause> represent the phrase and clause that reflect method or manner respectively, e.g., “using clustering algorithm” (verb phrase), “He solves the problems quickly” (adverb phrase) and “by using LexRank” (clause).

The following is the pattern of adverbial of purpose:

<adverbial of purpose> ::= <purpose clause>

<purpose clause> ::= <purpose subordinate conjunction> [<subject>] [<action>] [<object>] {<adverbial>} | “to” <action> [<object>] {<adverbial>}

<purpose subordinate conjunction> ::= “so”| “so that”| “so as to”| “in order that”| “in order to”| “for fear that”| “in case that”| “lest”| “for”

<purpose clause> represents the clause that reflects purpose, e.g., “to select the relevance sentences”.

The following is the pattern of adverbial of reason:

<adverbial of reason> ::= <reason clause>

<reason clause> ::= <reason subordinate conjunction> [<subject>] [<action>] [<object>] {<adverbial>}

<reason subordinate conjunction> ::= “because of”| “due to”| “owing to”| “as”| “since”| “because”| “based on”

<reason clause> represents the clause that reflects reason, e.g., “because the algorithm successfully reduces the complexity”.

The following is the pattern of adverbial of condition:

<adverbial of condition> ::= <condition clause>

<condition clause> ::= <condition subordinate conjunction> [<subject>] [<action>] [<object>] {<adverbial>}

<condition subordinate conjunction> ::= “if”| “unless”| “as long as”| “so long as”| “provided that”| “in case”| “in case of”| “on condition that”

<condition clause> represents the clause that reflects condition, e.g., “on condition that the algorithm can identify the wrong sentences”.

\section{Patterns of Question and Answer}
\subsection{Patterns of Question about Object and Patterns of Answer}
The question about object is used to determine the object of a sentence (consisting of thing and person).  The simple pattern of question about object is formed by 1) moving the object to the place in front of the subject of sentence (declarative sentence); 2) adding an auxiliary or modal verb to the place between the moved object and subject of sentence; and 3) adding an interrogative word to the place in front of the moved object.  

The following is a simple pattern of question about object.

<question about object> ::= {<adverbialQ>} <interrogative word> [<objectQ>] (<auxiliary>|<modal>) <subjectQ> <action> {<adverbialQ>}

<interrogative word> ::= “what” | “whom” | “which” | “whose”

<subjectQ> ::= <subject>

<actionQ> ::= [<auxiliary>|<modal>] <action>

<objectQ> ::= <noun phrase>

<adverbialQ> ::= <adverbial>

The following is an example of this pattern: “in search engine, what metadata does the database store?”, where “database” is <subjectQ>, “store” is <actionQ>, “metadata” <objectQ> and “in search engine” is <adverbialQ>.

The following is the answer pattern of the above question about object.

<answer of question about object> ::= {<adverbialA>} <subjectA> <actionA> <objectA> {<adverbialA>} 

<subjectA> ::= <subjectQ> | <subclass of <subjectQ>>

<actionA> ::= <actionQ> | <synonym of <actionQ>> | <subclass of <actionQ>> | <subclass of <actionA>>

<objectA> ::= <subclass of <objectQ>>

<adverbialA> ::= <adverbialQ> | <subclass of <adverbialQ>>

An example is: “database in master service stores metadata of the web pages in master-slave search engine”, which answers the question “in search engine, what does the database store?”, where “database in master service” is <subjectA> and it is the subclass of <subjectQ>, “store” is <actionA> and it is the same as <actionQ>, “metadata of the web pages” is <objectA> and it is the subclass of <objectQ> and it is the question asked for, and “in master-slave search engine” is <adverbialA> and it is the subclass of <adverbialQ>.

An object of a declarative sentence can be 1) single direct object, 2) direct object with indirect object, and 3) direct object with object complement, so the pattern of question about object needs to be extended as follows:

<question about direct object> ::= {<adverbialQ>} <interrogative word> [<direct objectQ>] (<auxiliary>|<modal>) <subjectQ> <action> [<preposition>] [<indirect objectQ>] {<adverbialQ>} | {<adverbialQ>} <interrogative word> [<direct objectQ>] (<auxiliary>|<modal>) <subjectQ> <action> <indirect objectQ> {<adverbialQ>} | {<adverbialQ>} <interrogative word> [<direct objectQ>] (<auxiliary>|<modal>) <subjectQ> <action> <object complementQ> {<adverbialQ>}

<interrogative word> ::= “what” | “whom” | “which” | “whose”

<subjectQ> ::= <subject>

<actionQ> ::= [<auxiliary>|<modal>] <action>

<direct objectQ> ::= <noun phrase>

<indirect objectQ> ::= <noun phrase>

<object complementQ> ::= <noun phrase>| <adjective phrase>

<adverbialQ> ::= <adverbial>

An example of this pattern is: “in unsupervised algorithm, which weight does the network send to node?”, where “network” is <subjectQ>, “send” is <actionQ>, “weight” is <direct objectQ>, “node” is <indirect objectQ> and “in unsupervised algorithm” is <adverbialQ>.

The following is the pattern of answering question about direct object.

<answer of question about direct object> ::= {<adverbialA>} <subjectA> <actionA> <direct objectA> [<preposition>] [<indirect objectA>] {<adverbialA>} | {<adverbialA>} <subjectA> <actionA> <indirect objectA> <direct objectA> {<adverbialA>} | {<adverbialA>} <subjectA> <actionA> <direct objectA> <object complementA> {<adverbialA>}

<subjectA> ::= <subjectQ> | <subclass of <subjectQ>>

<actionA> ::= <actionQ> | <synonym of <actionQ>> | <subclass of <actionQ>> | <subclass of <actionA>>

<direct objectA> ::= <subclass of <direct objectQ>>

<indirect objectA> ::= <indirect objectQ> | <subclass of <indirect objectQ>>

<object complementA> ::= <object complementQ> | <subclass of <object complementQ>>

<adverbialA> ::= <adverbialQ> | <subclass of <adverbialQ>>

An example is: “in graph-based unsupervised algorithm, the attention-based network iteratively sends adjacent nodes the updated weight”, which answers the question “in unsupervised algorithm, which weight does the network send to node?”, where “attention-based network” is <subjectA> and it is the subclass of <subjectQ>, “iteratively send” is <actionA> and it is the subclass of <actionQ>, “updated weight” is the <direct objectA> and it is the subclass of <direct objectQ> and it is the question asked for, “adjacent nodes” is the <indirect objectA> and it is the subclass of <indirect objectQ> and “in graph-based unsupervised algorithm” is <adverbialA> and it is the subclass of <adverbialQ>.

The following is the pattern of answering question about indirect object:

<question about indirect object> ::= {<adverbialQ>} <interrogative word> [<indirect objectQ>] (<auxiliary>|<modal>) <subjectQ> <action> <direct objectQ> <preposition> {<adverbialQ>} | {<adverbialQ>} <interrogative word> [<indirect objectQ>] (<auxiliary>|<modal>) <subjectQ> <action> <direct objectQ> {<adverbialQ>}

<interrogative word> ::= “what” | “whom” | “which” | “whose”

<subjectQ> ::= <subject>

<actionQ> ::= [<auxiliary>|<modal>] <action>

<direct objectQ> ::= <noun phrase>

<indirect objectQ> ::= <noun phrase>

<adverbialQ> ::= <adverbial>

An example of this pattern is “in unsupervised algorithm, which node does the network send the weight to?”, where “network” is <subjectQ>, “send” is <actionQ>, “weight” is <direct objectQ>, “node” is <indirect objectQ> and “in unsupervised algorithm” is <adverbialQ>.

The following is the pattern of answering question about indirect object:

<answer of question about indirect object> ::= {<adverbialA>} <subjectA> <actionA> <direct objectA> <preposition> <indirect objectA> {<adverbialA>} | {<adverbialA>} <subjectA> <actionA> <indirect objectA> <direct objectA> {<adverbialA>}

<subjectA> ::= <subjectQ> | <subclass of <subjectQ>>

<actionA> ::= <actionQ> | <synonym of <actionQ>> | <subclass of <actionQ>> | <subclass of <actionA>>

<direct objectA> ::= <direct objectQ> | <subclass of <direct objectQ>>

<indirect objectA> ::= <subclass of <indirect objectQ>>

<adverbialA> ::= <adverbialQ> | <subclass of <adverbialQ>>

An example of this pattern is “in graph-based unsupervised algorithm, the attention-based network iteratively sends the updated weight to the adjacent nodes”, which answers the questions “in unsupervised algorithm, which node does the network send the weight to?”, where “attention-based network” is <subjectA> and it is the subclass of <subjectQ>, “iteratively send” is <actionA> and it is the subclass of <actionQ>, “updated weight” is the <direct objectA> and it is the subclass of <direct objectQ> and “adjacent nodes” is the <indirect objectA> and it is the subclass of <indirect objectQ> and it is the question asked for.

The following is the pattern of answering question about object complement:

<question about object complement> ::= {<adverbialQ>} <interrogative word> [<object complementQ>] (<auxiliary>|<modal>) <subjectQ> <action> <direct objectQ> {<adverbialQ>}

<interrogative word> ::= “what” | “whom” | “which” | “whose”

<subjectQ> ::= <subject>

<actionQ> ::= [<auxiliary>|<modal>] <action>

<direct objectQ> ::= <noun phrase>

<object complementQ> ::= <noun phrase>| <adjective phrase>

<adverbialQ> ::= <adverbial>

An example of this pattern is “in computer science, what do researchers call deep learning?”, where “researchers” is <subjectQ>, “call” is <actionQ>, “deep learning” is <direct objectQ> and “in computer science” is <adverbialQ>.

The following is the pattern of answering question about object complement:

<answer of question about object complement> ::= {<adverbialA>} <subjectA> <actionA> <direct objectA> <object complementA> {<adverbialA>}

<subjectA> ::= <subjectQ> | <subclass of <subjectQ>>

<actionA> ::= <actionQ> | <synonym of <actionQ>> | <subclass of <actionQ>> | <subclass of <actionA>>

<direct objectA> ::= <direct objectQ> | <subclass of <direct objectQ>>

<object complementA> ::= <subclass of <object complementQ>>

<adverbialA> ::= <adverbialQ> | <subclass of <adverbialQ>>

An example of this pattern is “in computer science, senior researchers call deep learning representation learning” , which answers the question “in computer science, what do researchers call deep learning?”, where “senior researchers” is <subjectA> and it is the subclass of <subjectQ>, “call” is <actionA> which is the same as <actionQ>, “deep learning” is <direct objectA> and it is the same as <direct objectQ>, “in computer science” is <adverbialA> and it is the same as <adverbialQ> and “representation learning” is <object complementA> and it is the question asked for.

\subsection{Patterns of question about adverbial and answer}
The question about adverbial is used to know time, place, reason, purpose, method and condition of action.  The pattern of question about adverbial is formed by 1) adding an auxiliary or modal verb before the subject of sentence (declarative sentence), and 2) adding an interrogative word before the added auxiliary or modal verb of the sentence.

The following is the pattern of <question about adverbial>:

<question about adverbial> ::= {<adverbialQ>} <interrogative word> (<auxiliary>|<modal>) <subjectQ> <action> [<objectQ>] {<adverbialQ>} | {<adverbialQ>} <interrogative word> “to” <actionQ> <objectQ> {<adverbialQ>}

<interrogative word> ::= “when” | “where” | “why” | “how”

<subjectQ> ::= <subject>

<actionQ> ::= <action>

<objectQ> ::= <object>

<adverbialQ> ::= <adverbial>

An example of this pattern is: “When did Marie Curie win the Nobel Prize again?”, where “Marie Curie” is <subjectQ>, “win” is <actionQ>, “the Nobel Prize” is <objectQ> and “again” is the <adverbialQ>.

The following is the answer pattern of question about adverbial:

<answer of question about adverbial> ::= {<adverbialA>| <adverbialA*>} <subjectA> <actionA> [<objectA>] {<adverbialA>| <adverbialA*>}

<subjectA> ::= <subjectQ> | <subclass of <subjectQ>>

<actionA> ::= <actionQ> | <synonym of <actionQ>> | <subclass of <actionQ>> | <subclass of <actionA>>

<objectA> ::= <objectQ> | <subclass of <objectQ>>

<adverbialA> ::= <adverbialQ> | <subclass of <adverbialQ>>

<adverbialA*> ::= <adverbial of time> | <adverbial of place> | <adverbial of purpose> | <adverbial of reason> | <adverbial of method> | <adverbial of condition>

An example of this pattern is: “In 1911, Marie Curie won the Nobel Prize in chemistry again” , which answers the question “when did Marie Curie win the Nobel Prize again?”, where “Marie Curie” is <subjectA> and it is the same as <subjectQ>, “win” is <actionA> and it is the same as <actionQ>, “the Nobel Prize in chemistry” is <objectA> and it is the subclass of <objectQ>, “again” is the <adverbialA> and it is the same as <adverbialQ>, and “in 1911” is the <adverbial of time> and it is the question asked for.

\section{Comparison with Other Open Information Extraction Systems}
The difference between the common OpenIE systems (Stanford OpenIE, ClauseIE, MinIE and OllIE) and the proposed approach for extracting relation from the two sentences with complex representation is shown in TABLE D.1.  S1 is the sentence where the subject is an infinitive clause “to solve complex problems”, the Stanford OpenIE, ClauseIE and OllIE fail to extract relation from S1, MinIE extracts the right relation from S1.  S2 is the sentence where the subject is a clause with the lead word omitted “the experiment shows good results in China”. Stanford OpenIE misses the relation between “experiment” and “good results” in S2. ClauseIE classifies S2 to the wrong clause type, S2 is not the type of subject-action-object-complement but the type of subject-action-object.  MinIE and OllIE both miss the relation between “the experiment shows good results in China” and “researcher”.  In contrast, the proposed approach can correctly extract the relation (subject, action, object, and adverbial) in S1 and S2 by the pattern according to English grammar, and each syntax element is represented by a lexical structure.

\begin{table}
  \caption{Extracted Relations of Different OpenIE Systems.}
  \centering
  \begin{tabular}{>{\raggedright\arraybackslash}p{2cm} >{\raggedright\arraybackslash}p{6cm} >{\raggedright\arraybackslash}p{6cm} }
    \toprule
    \textbf{Systems} & \multicolumn{2}{>{\raggedright\arraybackslash}p{12cm}}{\textbf{Inputs}} \\
    \cmidrule(l){2-3}
    & S1: To solve complex problems is the essence of mathematics. & S2: The experiment shows good results in China encourages the researcher.\\
    \midrule
    Stanford OpenIE	& No extractions found.	& (“subject”: “good results”, “relation”: “is in”, “object”: “China”) \newline
(“subject”: “results”, “relation”: “encourages”, “object”: “researcher”)\newline
(“subject”: “good results”, “relation”: “encourages”, “object”: “researcher”) \\
    \midrule
    ClauseIE &	No extractions found.&	(SVOC, (“the experiment”, “shows”, “good results in China”, “encourages the researcher”))\newline
(SVO, (“the experiment”, “encourages”, “the researcher”)) \\
    \midrule
    MinIE &	(“to solve complex problem”, “is essence of”, “mathematics”) \newline
(“to solve complex problem”, “is”, “essence”) &	(“experiment”, “shows good results in”, “China”) \\
    \midrule
    OllIE&	No extractions found.&	(“the experiment”, “shows”, “good results”)\\
    \midrule
    The Proposed Approach&	“subject”: (clause, “lead word”: “to”, “subject”: Empty, “action”: (“”, “solve”, “”), “object”: (noun phrase, (“complex”, “problems”, “”)), “adverbial”: Empty), \newline
“action”: (“”, “is”, “”),\newline
“object”: (noun phrase, (“”, “essence”, “of mathematics”)),\newline
“adverbial”: Empty &	“subject”: (clause, “lead word”: Empty, “subject”: (noun phrase, (“”, “experiment”, “”)), “action”: (“”, “shows”, “”), “object”: (noun phrase, (“good”, “results”, “”)), “adverbial”: (adverbial of place, (“in”, (“”, “China”, “”)))),\newline
“action”: (“”, “encourages”, “”),\newline
“object”: (noun phrase, (“”, “researcher”, “”)),\newline
“adverbial”: Empty \\
    \bottomrule
  \end{tabular}
  \label{tab:table8}
\end{table}


\bibliographystyle{unsrt}  
\bibliography{references}

\end{document}